\documentclass[journal]{IEEEtran}
\usepackage{amsmath,amsfonts}
\usepackage{algorithmic}
\usepackage{algorithm}
\usepackage{dirtytalk}
\usepackage{array}
\usepackage[caption=false,font=normalsize,labelfont=sf,textfont=sf]{subfig}
\usepackage{textcomp}
\usepackage{stfloats}
\usepackage{url}
\usepackage{verbatim}
\usepackage{graphicx}
\usepackage{cite}
\usepackage{multirow}
\usepackage{array}
\usepackage{subcaption}
\usepackage[table]{xcolor}
\usepackage{mathrsfs}

\definecolor{lightred}{rgb}{1.0, 0.8, 0.8}
\definecolor{lightyellow}{rgb}{1.0, 1.0, 0.8}

\begin{document}

\title{Use of triplet loss for facial restoration in low-resolution images}

\author{Sebastián Pulgar, Domingo Mery 
\thanks{This work is supported byFondecyt-Chile 1191131 and National Center for Artificial Intelligence CENIA FB210017, Basal ANID, partly supported this work.}
}

\maketitle

\begin{abstract}
In recent years, facial recognition (FR) models have become the most widely used biometric tool, achieving impressive results on numerous datasets. However, inherent hardware challenges or shooting distances often result in low-resolution images, which significantly impact the performance of FR models. To address this issue, several solutions have been proposed, including super-resolution (SR) models that generate highly realistic faces. Despite these efforts, significant improvements in FR algorithms have not been achieved. In this paper, we propose a novel SR model called Face Triplet Loss GAN (FTLGAN), which focuses on generating high-resolution images that preserve individual identities rather than merely improving image quality, thereby maximizing the performance of FR models. The results are compelling, demonstrating a mean value of $d'$ 21\% above the best current state-of-the-art models, specifically having a value of $d' = 1.099$ and $AUC = 0.78$ for $14\times14$ pixels, $d' = 2.112$ and $AUC = 0.92$ for $28\times28$ pixels, and $d' = 3.049$ and $AUC = 0.98$ for $56\times56$ pixels.The contributions of this study are significant in several key areas. Firstly, a notable improvement in facial recognition performance has been achieved in low-resolution images, specifically at resolutions of $14\times14$, $28\times28$, and $56\times56$ pixels. Secondly, the enhancements demonstrated by FTLGAN show a consistent response across all resolutions, delivering outstanding performance uniformly, unlike other comparative models. Thirdly, an innovative approach has been implemented using triplet loss logic, enabling the training of the super-resolution model solely with real images, contrasting with current models, and expanding potential real-world applications. Lastly, this study introduces a novel model that specifically addresses the challenge of improving classification performance in facial recognition systems by integrating facial recognition quality as a loss during model training.

\end{abstract}

\begin{IEEEkeywords}
Face Recognition (FR), GAN, Triplet Loss, face re-identification.
\end{IEEEkeywords}

\section{Introduction} \label{intro}
\IEEEPARstart{I}{n} recent years, thanks to the emergence of artificial intelligence models, face recognition (FR) algorithms have achieved tremendous improvements that have led to the development of innovative FR models such as \cite{ada, arc, sphere, facenet}. These have achieved an accuracy of over 99\% on datasets such as LFW \cite{LFW}. These amazing results have turned facial recognition into the most widely used biometric technique in recent years, generating great contributions in areas such as security, finance, or even forensic cases \cite{du2021elements}.

However, despite the remarkable achievements made by neural models in the field of face recognition, their original design is oriented to high-resolution (HR) images, which hinders their direct application in contexts involving low-resolution (LR) images. This results in extensive problems in real applications such as surveillance, where capture distances and hardware limitations often result in very low-resolution facial images, with motion or even blurring, strongly decreasing the performance of these models \cite{Li_2019}.

\begin{figure}[!t]
\centering
\subfloat[]{\includegraphics[width=0.9in]{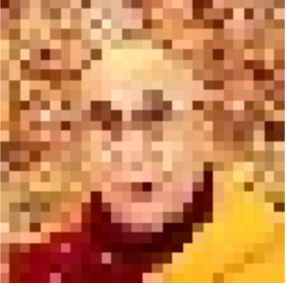}%
}
\subfloat[]{\includegraphics[width=0.9in]{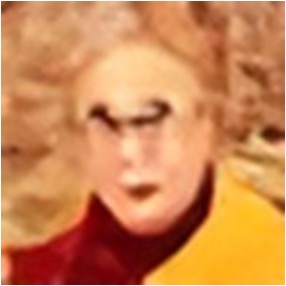}%
}
\subfloat[]{\includegraphics[width=0.9in]{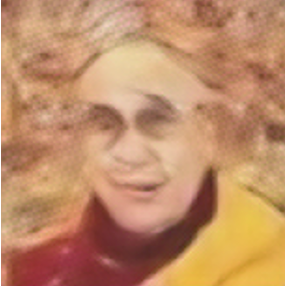}%
}

\caption{Visual comparison of three models in the process of super-resolution between 3 models, taking an image from $14\times14$ to $112\times112$ pixels. In a) results of a bicubic interpolation, in b) the results of the Real-SRGAN model \cite{wang2021realesrgantrainingrealworldblind}, and in c) the GFPGAN model.}
\label{fig_1}
\end{figure}

In the face of the severe performance degradation suffered by face recognition models with low-resolution images, two approaches have emerged: methods that learn a unified feature space and methods based on superresolution (SR) \cite{Li_2019}. Super-resolution models, which seek to generate a high-resolution (HR) face from a low-resolution (LR) input to improve face recognition, have presented a more successful approach in recent years compared to other methods \cite{9072532}. Despite the improvements presented by super-resolution mechanisms, there is still a wide challenge in very low-resolution images, such as $14\times14$, $28\times28$ and $56\times56$ pixels, where state-of-the-art models, such as GFPGAN or Real-SRGAN, generate visual deformations in faces (see Figure \ref{fig_1}) and loss of essential features that impede the re-identification task of FR algorithms. This fact has led bicubic interpolations to be considered as the best model in cases of very low-resolution faces since they preserve in a better way the original information of the face \cite{loreto_paper}.

The problems of deformations and poor face recognition (FR) performance affecting current super-resolution models are mainly attributed to excessive beautification of the restored images, leading to a loss of facial features during the face recognition process. This over-embellishment occurs because current models are trained with losses that prioritize the generation of realistic images, without taking into account a 
\newpage

\noindent variable that specifically evaluates and seeks to improve facial recognition. This variable, considered a second-order variable, is not included in the training of the models so its improvement is a second-order factor that can present improvements as a consequence of the beautification of the images.

Due to the problems that still exist in SR models, in this work, we will focus on the development of a low-resolution face restoration model that focuses on improving the face recognition process. This work has resulted in FTLGAN (Face Triplet Loss GAN) a novel super-resolution model that is able to maintain the identity of individuals by incorporating the quality of the FR as a loss in the generative network, which allows maintaining the distinctive features in very low-resolution cases. This model will be compared with several models evaluated in \cite{loreto_paper,loreto} at the resolutions of $14\times14$, $28\times28$ and $56\times56$ pixels, in the VGG-Face 2 dataset \cite{vggface_2}, following the protocols and performing a fair comparison between the models.

The contributions of the present work are significant in several key areas. Firstly, a notable improvement in facial recognition performance has been achieved in low-resolution images, specifically at resolutions of $14\times14$, $28\times28$, and $56\times56$ pixels. Secondly, the experienced enhancements by the model demonstrate a consistent response across all resolutions, consistently delivering outstanding performance, unlike other comparative models. As a third contribution, an innovative approach has been implemented using triplet loss logic, enabling the training of a super-resolution model solely with real images, contrasting with current models, thus expanding the potential for real-world applications. Lastly, as a fourth contribution, a novel model has been introduced that specifically addresses the challenge of improving classification performance in facial recognition systems by integrating facial recognition quality as a loss during model training.

The next of the paper is organized as follows. In section 2, a detailed literature review will address the theoretical basis of the FTLGAN model, including aspects such as the face-reidentification process, face-recognition models, super-resolution model, and relevant evaluation metrics. In addition, the dataset used for the study will be presented and its relevance will be discussed. In section 3 will be devoted to the detailed exposition of the FTLGAN architecture, going in-depth into its components and its operation. Subsequently, in section 4, experimental results of the FTLGAN model will be provided, accompanied by a detailed ablation analysis to better understand its performance. In section 5 will be devoted to an in-depth discussion of the results obtained, analyzing their implications and possible limitations. Finally, in section 6 will present the conclusions derived from this study, highlighting key findings and possible future directions for research.

\section{RELATED WORK AND DATASETS}
This section will present the different face recognition techniques, the different super-resolution mechanisms, and the dataset used to evaluate the performance of the different models.

\subsection{Face re-identification}

\begin{figure}[t!]
            \centering
            \includegraphics[width=0.95\linewidth]{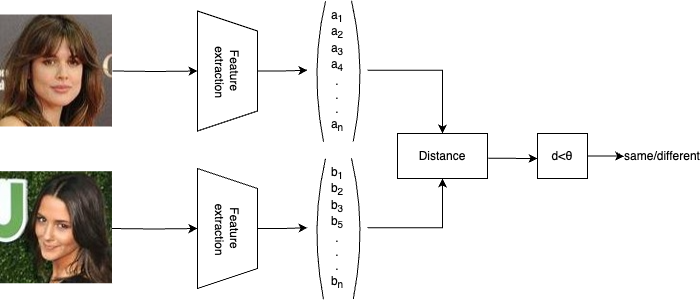}
            \caption{Explanatory diagram of the face re-identification task: an example of two faces that do not belong to the same identity is presented. In this case $d < \theta$.}
            \label{gen0}
        \end{figure}

Facial re-identification is the process of determining whether two facial images captured at different times and with different cameras represent the same person \cite{Parkhi2015,Luo_Zhu_Liu_Wang_Tang_2016 }. Unlike other facial tasks, such as facial verification, which verifies whether a given face corresponds to a specific person, and facial recognition, which identifies a person from a given image, facial re-identification involves comparing facial features to establish the similarity between two images, as can be seen in the Figure \ref{gen0}.

This process becomes particularly challenging when one of the images is low resolution (LR), as the reduced quality can make it difficult to accurately extract facial features \cite{Cheng_2020_CVPR, 8985713}. Low-resolution person re-identification is an important area of research, especially in surveillance and security applications, where images captured by cameras are often of low quality due to factors such as distance, viewing angle, and variable illumination.

\subsection{Face Recognition Models (FR)}

To make consistent comparisons between faces, face recognition algorithms are used. These are mechanisms capable of extracting the characteristics of the facial images through an n-dimensional vector, obtaining a numerical representation of the faces, which allows comparing the similarity or distance between the images \cite{Kortli2020}.

There are numerous methodologies to perform feature extraction, however, deep learning has become the predominant mechanism in the last ten years, allowing the development of a multitude of models, which are characterized by using similar backbones, but different loss functions. Depending on the type of loss function, FR algorithms can be classified into three major groups \cite{Wang2021}:

\subsubsection{Euclidean-distance-based loss}

Models based on Euclidean distance are characterized by using vector representations of the faces to incorporate them into the Euclidean space, seeking to reduce the intravariance and increase the intervariance between faces. The most popular loss using this principle is contrastive loss, a type of loss that seeks to minimize the Euclidean distance between positive face representations (same person) and maximize the Euclidean distance between negative face representations (different persons) \cite{sun2015deeply}, by the following equation:

\begin{multline}
    \mathscr{L} =y_{i j} \max \left(0,\left\|f\left(x_i\right)-f\left(x_j\right)\right\|_2-\epsilon^{+}\right) \\ 
      +\left(1-y_{i j}\right) \max \left(0, \epsilon^{-}-\left\|f\left(x_i\right)-f\left(x_j\right)\right\|_2\right) \,,
\end{multline}

\noindent where $y_{i j}=1$ means $x_i$ and $x_j$ are matching samples and $y_{i j}=0$ means non-matching samples. $f(\cdot)$ is the feature embedding, and $\epsilon^{+}$and $\epsilon^{-}$control the margins of the matching and non-matching pairs respectively.

Among the face recognition models that use Euclidean losses, FaceNet \cite{facenet} stands out, which uses a triplet loss type loss, which, unlike the constraint loss, which takes into account the absolute distances of matched and mismatched pairs, considers the relative difference between them, according to the next formula:

\begin{equation}\label{tripletloss}
    \left\|f\left(x_i^a\right)-f\left(x_i^p\right)\right\|_2^2+\alpha<\left\|f\left(x_i^a\right)-f\left(x_i^n\right)\right\|_2^2 \,, 
\end{equation}

\noindent where $x_i^a, x_i^p$ and $x_i^n$ are the anchor, positive and negative samples, respectively, $\alpha$ is a margin and $f(\cdot)$ represents a nonlinear transformation embedding an image into a feature space.

\subsubsection{Angular/cosine-margin-based loss}
The angular losses arise from the Softmax loss concept which is characterized by training focused on classifying faces into classes representing identities, which presents serious inter/itra-class problems \cite{Wang2021}. To improve the problems of the Softmax model, the use of angular/cosine margin-based loss was proposed, generating a margin between classes that are located in the cortex of a hypersphere allowing for better classification \cite{liu2016large}.

Angular models are based on the intrinsic angular behavior of the softmax loss, located in the crust of a feature hypersphere, reformulating the softmax expression as a function of the angle between the feature vector and the column vector of weights, which allows the emergence of state-of-the-art models, such as ArcFace \cite{arc} and AdaFace \cite{ada}. These models, in addition to expressing the function in terms of the angle, incorporate a margin, allowing better differentiation:

\begin{equation}\label{eq_arcface}
    \scriptstyle
    L=-\frac{1}{N} \sum_{i=1}^N \ln \frac{\exp \left\{s \cdot \cos \left(\theta_{y_i, i}+m\right)\right\}}{\exp \left\{s \cdot \cos \left(\theta_{y_i, i}+m\right)\right\}+\sum_{j \neq y_i} \exp \left\{s \cdot\left(\cos \left(\theta_{j, i}\right)\right\}\right.} \,.
\end{equation}
\vspace{2mm}

\subsubsection{Loss variations}
Numerous studies have proposed variations on the softmax and angular models by generating normalizations of the characteristics and weights of the loss functions to improve the performance of the models \cite{liu2016large} as follows:

\begin{equation}\label{ec000001}
    \widehat{W}=\frac{W}{\|W\|}, \widehat{x}=\alpha \frac{x}{\|x\|} \,,
\end{equation}

\noindent where $\alpha$ is a scalar parameter, $x$ is the learned feature vector and $W$ are the weights of the last fully connected layer.

\subsection{Upsampling methods}
Super-resolution mechanisms are responsible for converting low-resolution (LR) images into high-resolution (HR) images, seeking to preserve as much detail of the identity of persons as possible in the case of facial images. This process of transforming low-resolution images to high-resolution is known as upsampling operation and can be divided into two types: interpolation methods and learning-based upsampling \cite{li2021beginner}.

\subsubsection{Interpolation methods}
Interpolation is the most commonly used oversampling method \cite{li2021beginner}. The interpolation-based upsampling methods performs a scaling only using information from known pixels to estimate the value of unknown pixels, being an easy-to-implement methodology \cite{li2021beginner}. This logic has allowed the emergence of several subtypes of interpolations among which stand out: 

\begin{itemize}
    \item \textbf{Nearest-neighbor Interpolation:} a model that selects the nearest pixel value for each position to be interpolated independently of any other pixel.

    \item \textbf{Bilinear Interpolation:} a model that performs linear interpolation on one axis of the image and then performs it on the other axis.

    \item \textbf{Bicubic interpolation:} which similarly to Bilinear interpolation performs a cubic interpolation on each of the two axes, however, it takes into account 4$\times$4 pixels and produces smoother results with fewer artifacts \cite{fadnavis2014image}.
    
\end{itemize}

\subsubsection{Learning-based Upsampling} \label{cosax}
Unlike interpolation models, learning base upsample models are characterized by learning end-to-end resampling by intruding learning convolutional layers. Among these, two logics stand out.

\begin{itemize}
    \item \textbf{Sub-pixel Convolutional Layer:} Also known as the deconvolution layer, it is responsible for performing an inverse transformation to the standard convolution. Its main purpose is to predict the possible input from feature maps that have a similar dimension to the convolution output. In essence, this layer seeks to increase the resolution of the image through an expansion process involving the insertion of zeros, followed by the application of the convolution operation \cite{Zeiler2014}.

    \item \textbf{Sub-pixel Convolutional Layer:} This corresponds to another mechanism that is fully learnable end-to-end and performs upsampling by generating multiple channels through convolution and subsequent reshaping. Within this layer, an initial convolution is implemented to produce outputs with $s^2$ times the channels, where $s$ denotes the scale factor. This convolution process is repeated within the layer to generate outputs with $s^2$ times the channels, where $s$ continues to represent the scale factor \cite{Shi_2016_CVPR}.

\end{itemize}

\subsection{Evaluation metrics}

\begin{figure}[t!]
        \centering
        \includegraphics[width=0.9\linewidth]{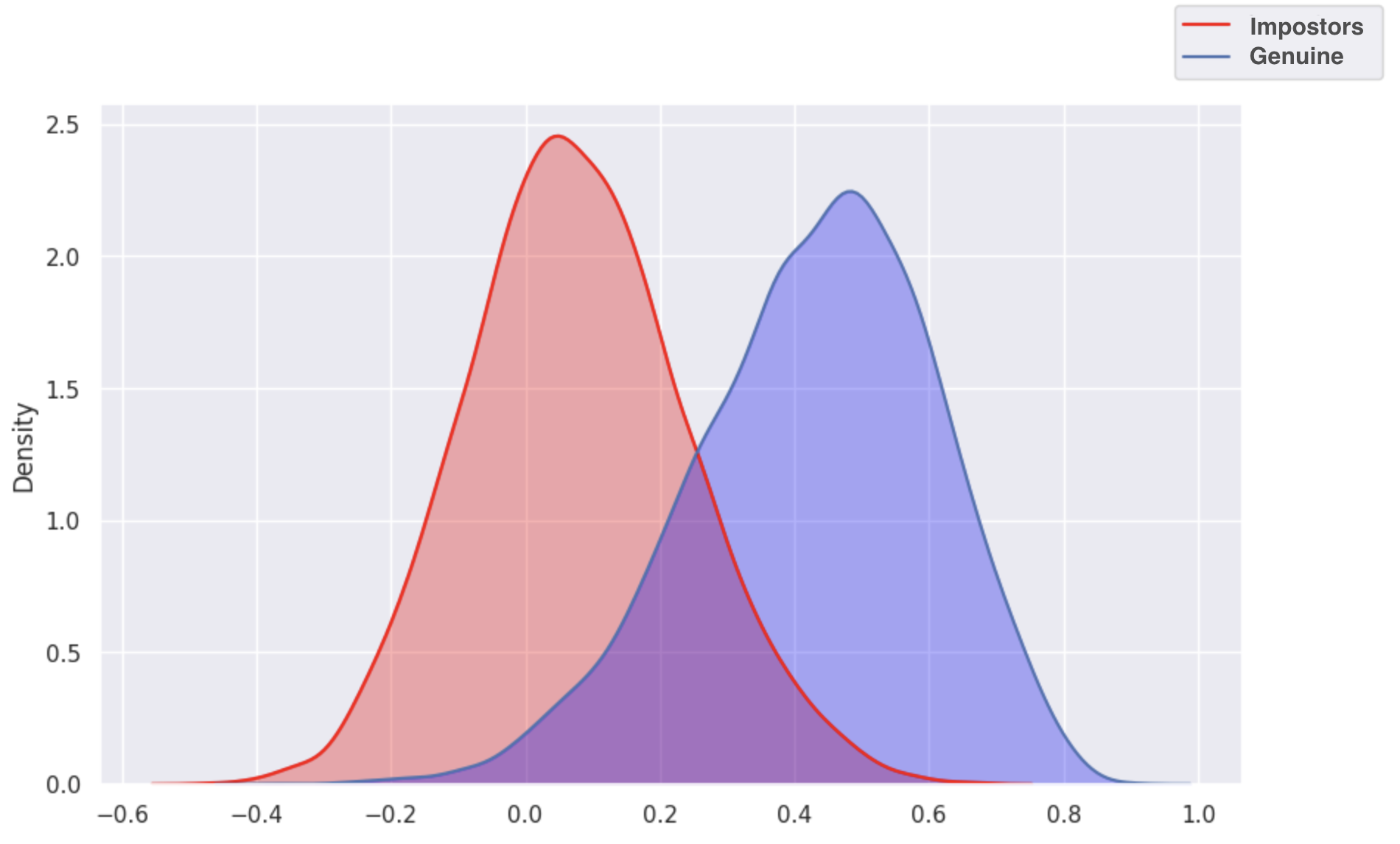}
        \caption{Example of a curve of genuine and impostors, using Euclidean distance between pairs, where the two curves should be as far apart as possible to improve task performance.}
        \label{gen}
    \end{figure}

Once the vectors and the distance between the pairs have been calculated, it is possible to evaluate the performance of the model. For this purpose, tests are performed on all the pairs of the dataset, and the performance of the model is evaluated, visualizing that the pairs of impostors (faces of different persons) are recognized as different persons and the pairs of genuines (faces of the same person) are effectively recognized as the same identity. Since the model performance is variable depending on the selected threshold, a genuine and impostor curve is performed in which all dataset pairs are evaluated for all possible thresholds, generating two curves showing the classification status, as can be seen in the example in Figure \ref{gen}.

    To evaluate objectively (and not visually) the separation between the curves and the confusion zone, parameter $d'$ is calculated as a metric using the:

    \begin{equation}\label{d_prim}
        d^{\prime}(g, i) = \frac{|\mu_g - \mu_i|}{\sqrt{\frac{\sigma_g^2 + \sigma_i^2}{2}}}
    \end{equation}

    This parameter allows a less ambiguous comparison, considering the mean and the standard deviation of the curves.

\begin{figure}[t!]
        \centering
        \includegraphics[width=0.9\linewidth]{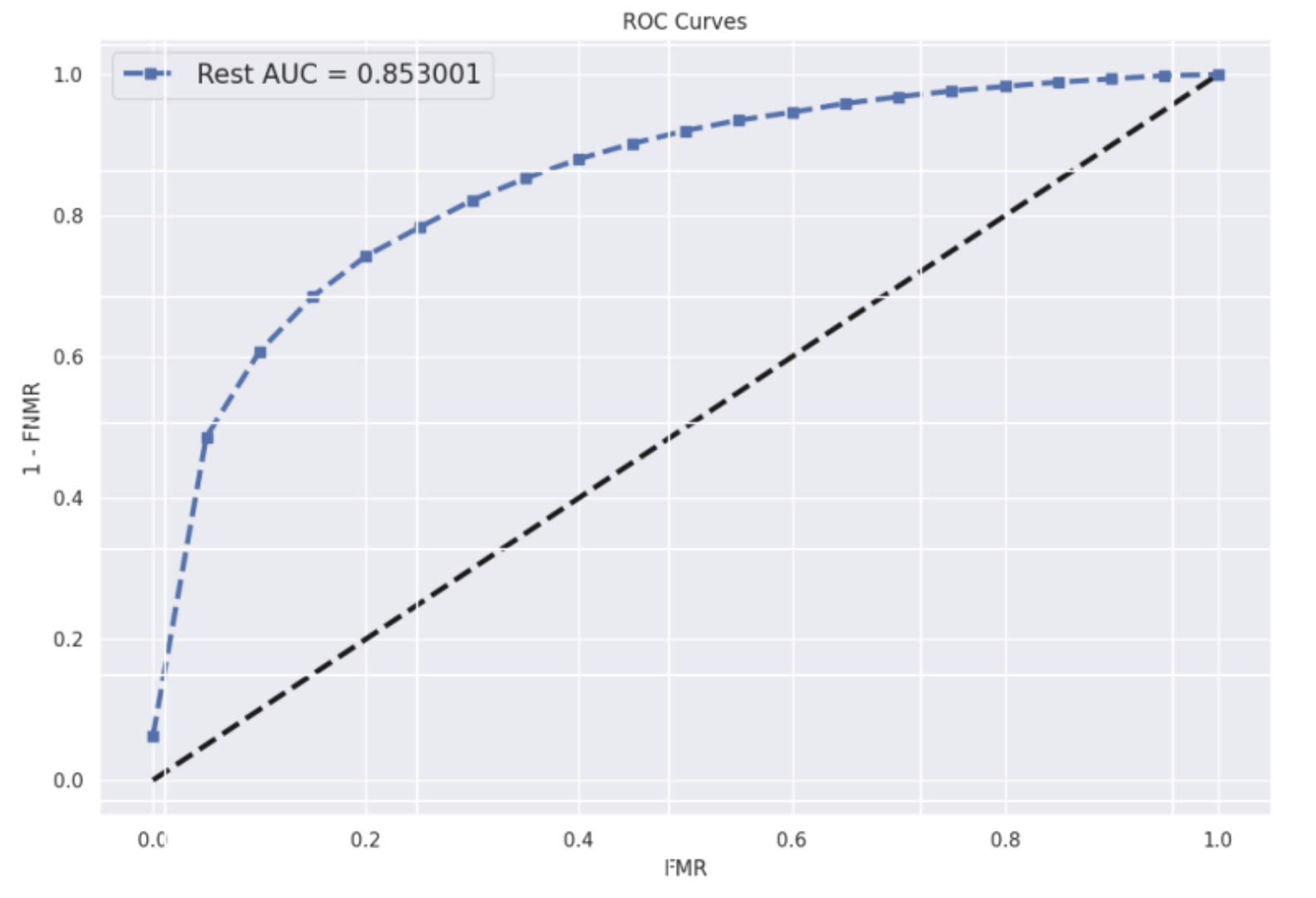}
        \caption{Example of a roc where the FMR can be seen on the x-axis and 1-FNMR on the y-axis. The blue line corresponds to the ROC curve while the red curve corresponds to an ROC with AUC 0.5 showing randomness or total confusion between classes.}
        \label{roc_realx}
    \end{figure}

     On the other hand, a Receiver Operating Characteristic (ROC) \cite{Fawcett2006} curve is plotted, which allows to verify the performance of the model for all thresholds by means of a graph of FMR vs FNMR as shown in Figure \ref{roc_realx}.

\section{Proposed method: FTLGAN}

Due to the mentioned problems with the super-resolution models, this section will first, explore the development of the FTLGAN model, followed by the presentation of several variants of the FTLGAN model.

\subsection{Triplet Loss Training}
As mentioned in the introduction, current super-resolution models often do not incorporate the embeddings of face recognition models in their training. In this context, the FTLGAN model focuses on rethinking the traditional training logic of GANs and face enhancement models. To achieve this, the FTLGAN consists of two stages in its training process, which operate together following a triplet loss-based logic. During training, image triplets composed of a low-resolution $Anchor$ image (the target identity), a high-resolution $Positive$ image (the same target identity), and a high-resolution $Negative$ face (of another identity) are used.

\begin{figure}[!t]
\centering
\includegraphics[width=3.1in]{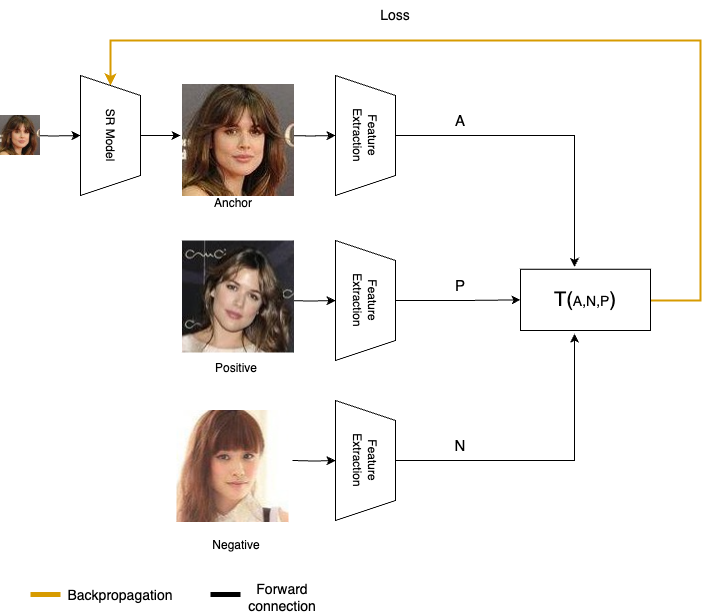}
\caption{Triplet loss training architecture applied to the superresolution process, where T(A,N,P) corresponds to the triplet loss function applied on the vector of characteristics generated for the anchor image, negative and positive correspondingly.}
\label{figlogica}
\end{figure}

In the first stage, called \say{generative}, a neural network acts as a decoder performing the scaling process on the low-resolution $Anchor$ image, converting it into a high-resolution image of equal size to the $Positive$ and $Negative$ images. In the second stage, called \say{feature extraction}, a pre-trained face recognition algorithm, with weights frozen, is used to extract a latent vector of the $Positive$, $Negative$, and $Anchor$ face with scaling. The quality of the image restored by the decoder is evaluated by calculating the triplet loss between the latent vectors, and backpropagation is performed to train the generative decoder, as shown in Figure \ref{figlogica}.

In order to focus the results on the training logic instead of the blocks used, we decide to use as a decoder the ESGAN model generator \cite{wang2018esrgan}, a widely recognized architecture that differs from other current GANs, such as styleGAN \cite{karras2019stylebased}, by having superresolution tasks as its main objective. The topology used was identical to the one proposed in \cite{wang2018esrgan}, using 16 RRDB residual blocks. This allows us to have a versatile model targeting various tasks, including image quality enhancement in the SR task \cite{Lim_2017_CVPR_Workshops}.

In the feature extraction phase, we propose to integrate a feature extraction model trained with triplet loss logic. In this line, the FaceNet \cite{facenet} architecture was selected. The FaceNet architecture implemented for the FTLGAN model employs a ResNet100 backbone \cite{he2015deep}, previously trained on the VGG-Face 2 dataset \cite{vggface_2}.

In addition to allowing a focus on face recognition quality, the triplet loss training logic can be trained with real low-resolution images, since no comparison between the restored image and an ideal image is required. This mitigates a problem present in \cite{wang2018esrgan,wang2021realworld,yang2021gan}, which need to train with synthetic low-resolution images generated from high-resolution image compressions. This effect allows the FTLGAN model to learn effects other than low resolution, including blurring and noising.

\subsection{Perceptual Loss}

In our approach, it is crucial to note that the proposed model does not present any loss that works directly in the image space and controls that the result actually looks like a face. Instead, the model seeks to optimize the n-dimensional representations of the faces which indirectly impacts on obtaining realistic images. This second-order strategy implies that, during the first few training epochs, the model may experience noticeable divergence due to the complexity of learning the subtle, nonlinear correlations that characterize facial features. By not imposing strict constraints from the outset, the model has the flexibility to adapt and adjust to the inherent diversity in facial appearance, although this initial process may result in less accurate or consistent results. To mitigate potential divergence in early epochs and guide the model toward generating more consistent facial images, a second loss is known as the perceptual loss \cite{johnson2016perceptual}.

The perceptual loss proposed by Johnson et al. \cite{Johnson2016} is based on the concept of minimizing the distance between the features activated in a reference image and another restored in a deep network based on the idea of being closer to the perceptual similarity \cite{NIPS2015_a5e00132}. In the case of FTLGAN, perceptual loss is implemented in a VGG19-54 network following the architecture defined in \cite{wang2021realesrgan}. In this, a 19-layer pre-trained VGG \cite{simonyan2015deep} network is used, where \say{54} indicates the features obtained from the 4th convolution before the 5th max-pooling layer. By using the 4th layer, we can capture features that are not too deep while maintaining strong supervision.

\begin{figure}[!t]
\centering
\includegraphics[width=3.3in]{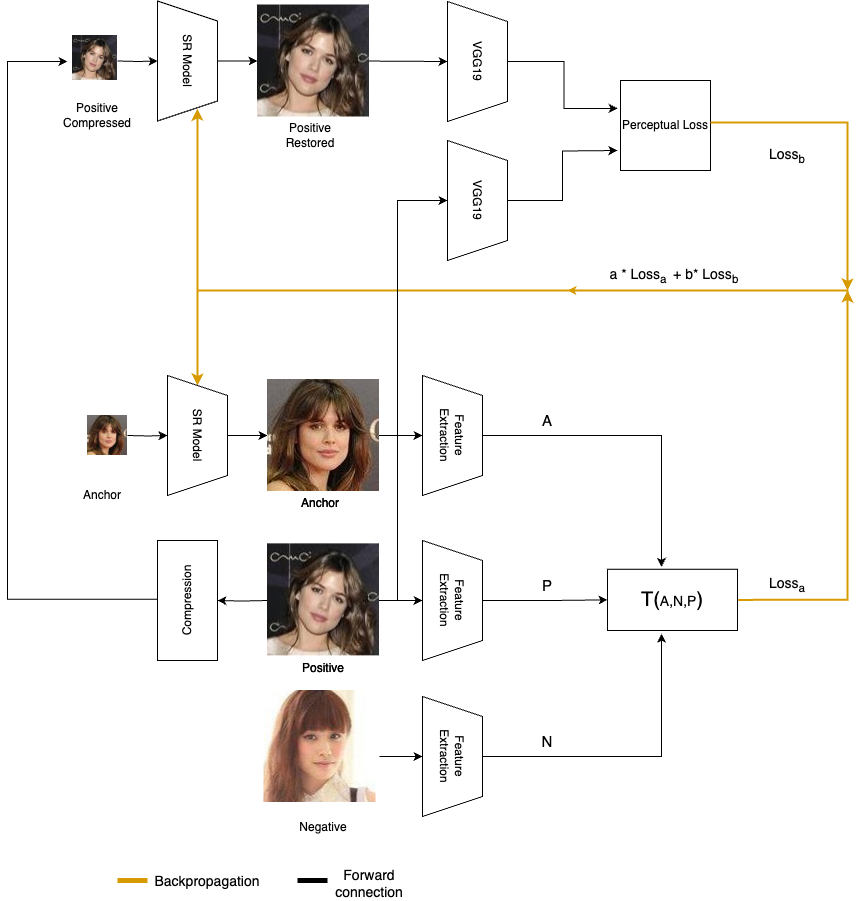}
\caption{FTLGAN training architecture using in one branch training triplets in conjunction with a perceptual loss, it is possible to visualize how the positive image of the triplet is compressed to obtain the perceptual loss.}
\label{arq}
\end{figure}

Although the incorporation of the perceptual loss can help the model by integrating more direct relations that allow for the reduction of the divergence in the first epochs, this topology presents difficulties in its implementation. It prevents training with only real images of low resolution since both a low and high-resolution image are required to perform the training supervision. To mitigate this drawback, it was proposed to perform a synthetic compression of the $Positive$ image using a bicubic interpolation, generating the architecture that can be visualized in figure \ref{arq}.

Thus the use of $L_{percep}$ and the $L_{triplet}$ results in an overall loss for the FTLGAN model defined by the linear combination between both losses in the form:

\begin{equation}\label{triplet_loss_eq}
    L = \alpha L_{percep} + \beta L_{triplet},
\end{equation}

For this work, we tested other losses that incorporate more direct correlations, such as MSE \cite{kato2021mse}, in conjunction with Triplet Loss. However, no other mechanism presented better results than the presented combination.

\section{Experiments}

In order to obtain comparable and valid results, the proposed FTLGAN model was evaluated following the experimental protocol proposed in \cite{loreto} in which $14\times14$, $28\times28$ and $56\times56$, scaling to $\times$8, $\times$4 and $\times$2 to generate high-resolution images of 112 pixels.

\subsection{Dataset}

In the world of face recognition, finding complete datasets that allow a deep and comparable evaluation of the data is a challenge, in this line to achieve replicable and comparable results, it was decided to use the dataset proposed in \cite{loreto}, generating with this comparable results with other SR models. This Dataset is conformed by an edited version of VGG-Face 2 \cite{vggface_2}, which contains high-resolution images of $112\times112$ pixels and low-resolution images of three types of resolution: $56\times56$, $28\times28$ and $14\times14$ pixels.

Each group of the dataset is made up of a total of 163,564 training images with 8,605 different identities, in addition to 8,791 test images with 497 different identities. It is important to note that in the creation of these sets, no enlargement was performed, so no new information was created, however, in some cases given the protocol, minimal downscaling had to be performed to standardize the dimensions, for which bicubic interpolation was used.

In addition, to the images in the four resolution types, the dataset has 163,564 triplets of data for each resolution in the training set in which an LR anchor image, a positive HR image (same identity), and a negative HR image (different identity) are presented to enable contrastive or triple-loss type training.

\subsection{Training details and parameters}
For training the FTLGAN model, a linear combination between perceptual loss and triplet loss was used. In all experiments, we set $\alpha = 0.8$ and $\beta = 0.2$, following the equation \ref{triplet_loss_eq}, which intensified the impact of triplet loss on perceptual loss. In addition, a learning rate of $1e^{-5}$ was used for all experiments.

The burial process was performed independently for each of the dataset resolutions ($14\times14$, $28\times28$, and $56\times56$). Ten epochs on an NVIDIA RTX3090 Ti video card were used for each of the training.

\subsection{Experiments results}

The results obtained from the FTLGAN model are compared with both the interpolation methods and with learning-base upsampling methods shown in section \ref{cosax}, generating the Table \ref{tab:final}. When observing the results it is possible to notice that the proposed FTLGAN model presents the best results in $28\times28$ and $56\times56$ resolution, as well as a better average than all the other models SR presented in \cite{loreto_paper}.

\begin{table*}[!t]
\caption{Comparative table of $d'$ and $AUC$ values between the state-of-the-art restoration-based learning and interpolation upsampling models, presented in Prieto (2022), and the FTLGAN model, for resolutions of $14\times14$, $28\times28$ and $56\times56$.}
\centering
{\renewcommand{\arraystretch}{1.45}
\resizebox{0.88 \textwidth}{!}{ 
\begin{tabular}{|c|c|c|c|c|c|c|c|c|c|}
\hline & & \multicolumn{2}{|c|}{$\mathbf{1 4 x 1 4}$} & \multicolumn{2}{c|}{$\mathbf{2 8 x 2 8}$} & \multicolumn{2}{c|}{$\mathbf{5 6 x 5 6}$} & \multicolumn{2}{c|}{ Average } \\
\cline { 2 - 10 } {} & \textbf{Exp. name} & $\mathbf{d}^{\prime}$ & AUC & $\mathbf{d}^{\prime}$ & AUC & $\mathbf{d}^{\prime}$ & AUC & $\mathbf{d}^{\prime}$ & AUC \\
\hline \multirow{13}{*}{\rotatebox[origin=c]{90}{\textbf{Conventional methods}}} & Baseline \cite{loreto}& 0.411 & 0.61 & 0.933 & 0.74 & 1.523 & 0.86 & 0.956 & 0.74 \\
\cline { 2 - 10 }  & Bicubic + Facenet \cite{loreto}& 0.462 & 0.63 & 1.9 & 0.91 & 2.787 & 0.97 & 1.716 & 0.84 \\
\cline { 2 - 10 } & Bicubic + AdaFace \cite{loreto}& 0.369 & 0.60 & 1.715 & 0.88 & 2.528 & 0.95 & 1.537 & 0.81 \\
\cline { 2 - 10 }  & T. GAN + Arcface \cite{loreto}& 0.253 & 0.57 & 0.959 & 0.75 & 1.547 & 0.87 & 0.92 & 0.73 \\
\cline { 2 - 10 } & T. GAN + GAN T. S. \cite{loreto}& 1.156 & 0.79 & 1.421 & 0.84 & 1.388 & 0.83 & 1.322 & 0.82 \\
\cline { 2 - 10 }  & Area + GAN T. S. \cite{loreto}& 0.448 & 0.62 & 0.582 & 0.66 & 0.653 & 0.67 & 0.561 & 0.65 \\
\cline { 2 - 10 }  & Bicubic + GAN T. S. \cite{loreto}& 0.619 & 0.67 & 0.724 & 0.69 & 0.72 & 0.69 & 0.688 & 0.68 \\
\cline { 2 - 10 } & Lanc. + GAN T. S. \cite{loreto}& 0.613 & 0.67 & 0.704 & 0.69 & 0.708 & 0.69 & 0.675 & 0.68 \\
\cline { 2 - 10 }  & Nearest + GAN T. S. \cite{loreto}& 0.448 & 0.62 & 0.582 & 0.66 & 0.653 & 0.67 & 0.561 & 0.65 \\
\cline { 2 - 10 }  & Area + Area T. S. \cite{loreto}& 1.202 & 0.80 & 1.40 & 0.84 & 1.42 & 0.84 & 1.341 & 0.83 \\
\cline { 2 - 10 }  & Bicubic + Bicubic T. S. \cite{loreto}& 1.041 & 0.76 & 1.373 & 0.83 & 1.424 & 0.84 & 1.279 & 0.81 \\
\cline { 2 - 10 }  & Lanc. + Lanc. T. S. \cite{loreto}& 1.007 & 0.76 & 1.371 & 0.83 & 1.426 & 0.84 & 1.268 & 0.81 \\
\cline { 2 - 10 }  & Nearest Nearest T. S. \cite{loreto}& \textbf{1.236} & \textbf{0.81} & 1.416 & 0.84 & 1.449 & 0.84 & 1.367 & 0.83 \\
\hline \multirow{6}{*}{\rotatebox[origin=c]{90}{\textbf{Specialized GAN}}}  & ESRGAN + GAN T.S. \cite{loreto_paper}& 0.514 & 0.64 & 0.894 & 0.73 & 1.106 & 0.78 & 0.838 & 0.72 \\
\cline { 2 - 10 }  & GPEN (256) + ARCFACE \cite{loreto_paper}& 0.31 & 0.57 & 0.973 & 0.76 & 1.46 & 0.76 & 0.914 & 0.7 \\
\cline { 2 - 10 }  & GPEN (512) + ARCFACE \cite{loreto_paper}& 0.261 & 0.54 & 1.121 & 0.74 & 1.632 & 0.79 & 1.005 & 0.69 \\
\cline { 2 - 10 }  & GFPGAN (V1) + ARCFACE \cite{loreto_paper}& 0.232 & 0.54 & 0.945 & 0.71 & 1.553 & 0.84 & 0.91 & 0.7 \\
\cline { 2 - 10 }  & GFPGAN (V2) + ARCFACE \cite{loreto_paper}& 0.203 & 0.53 & 0.951 & 0.72 & 1.603 & 0.85 & 0.919 & 0.7 \\
\cline { 2 - 10 }  & ESRGAN + GFPGAN(V2) \cite{loreto_paper}& 0.213 & 0.53 & 1.453 & 0.84 & 1.763 & 0.88 & 1.143 & 0.75 \\
\hline \rotatebox[origin=c]{90}{\textbf{ Ours }}  & \textbf{FTLGAN +FaceNet} & 1.099 & 0.78 & \textbf{2.112} & \textbf{0.92} & \textbf{3.049} & \textbf{0.98} & \textbf{2.086} & \textbf{0.89} \\
\hline
\end{tabular}}}
\label{tab:final}
\end{table*}

When comparing the data, it is possible to notice a clear improvement of the FTLGAN model concerning all the other topologies presenting a performance 11\% better in $28\times28$, 9.4\% better in $56\times56$ and 21\% better on average than the best model of the state of the art and baseline in this problem, only being surpassed by Nearest + Nearest T. S in $14\times14$ pixels.

\begin{figure*}
    \centering
    \captionsetup[subfigure]{labelformat=empty}
    \includegraphics[width=0.32\linewidth]{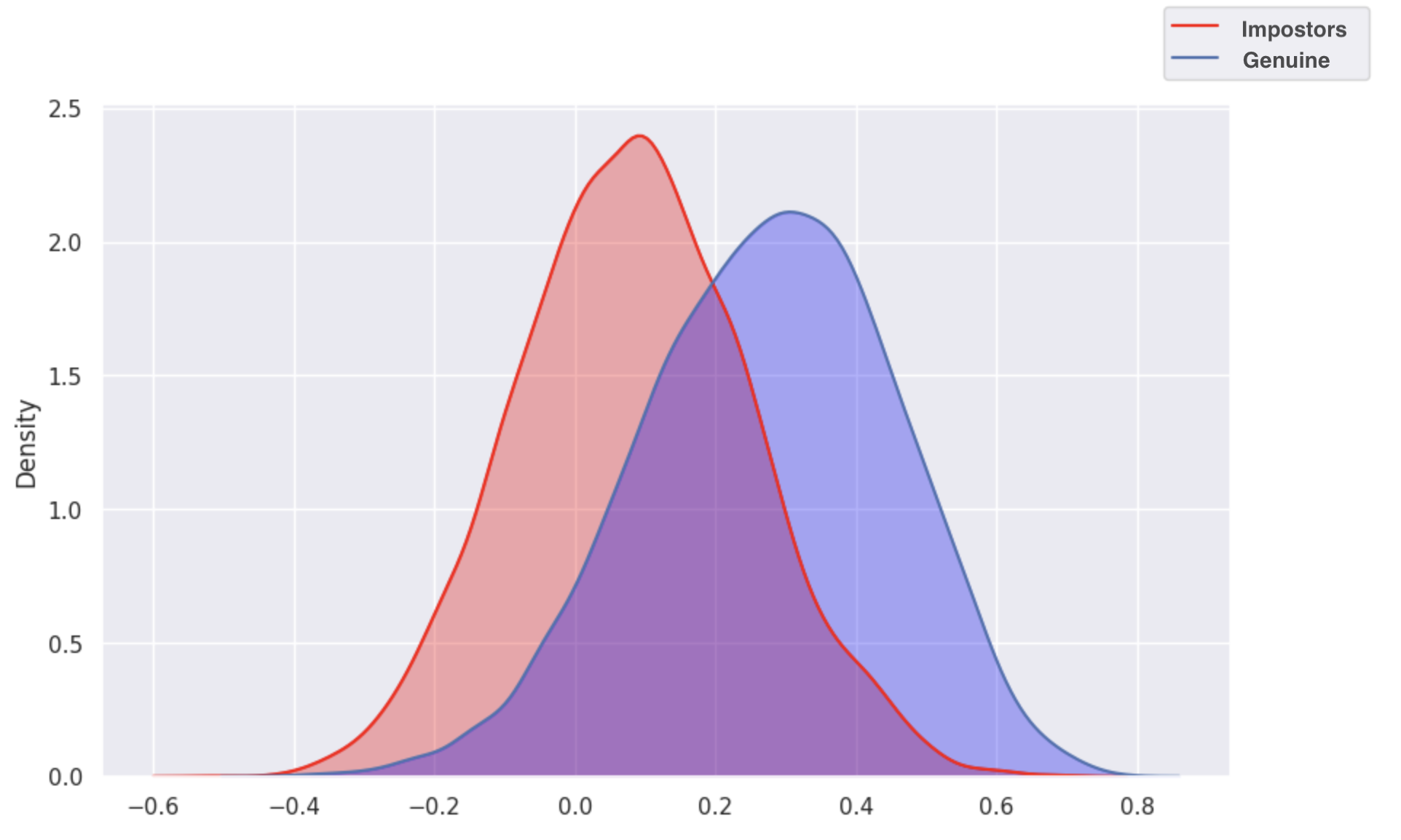}
    \includegraphics[width=0.32\linewidth]{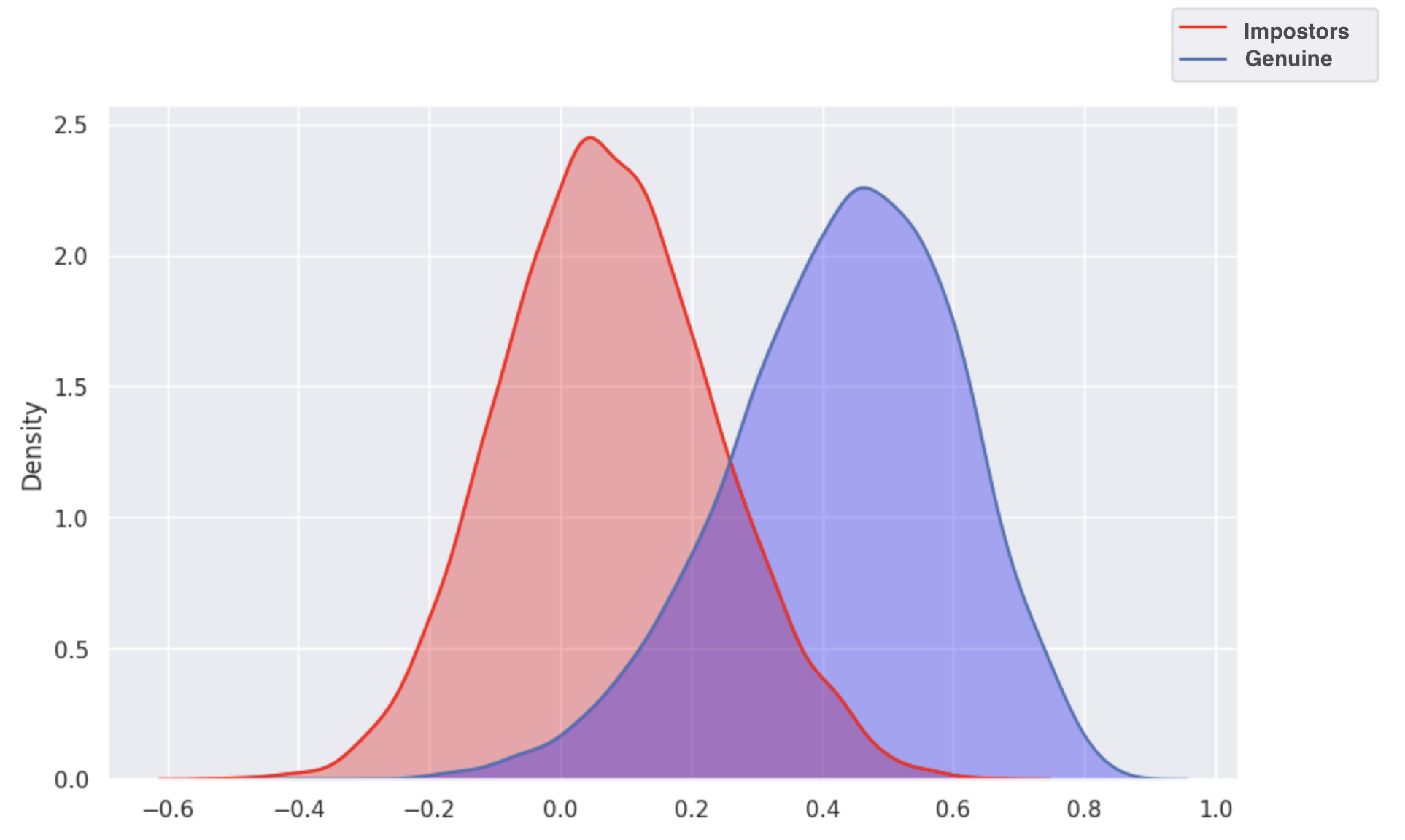}
    \includegraphics[width=0.32\linewidth]{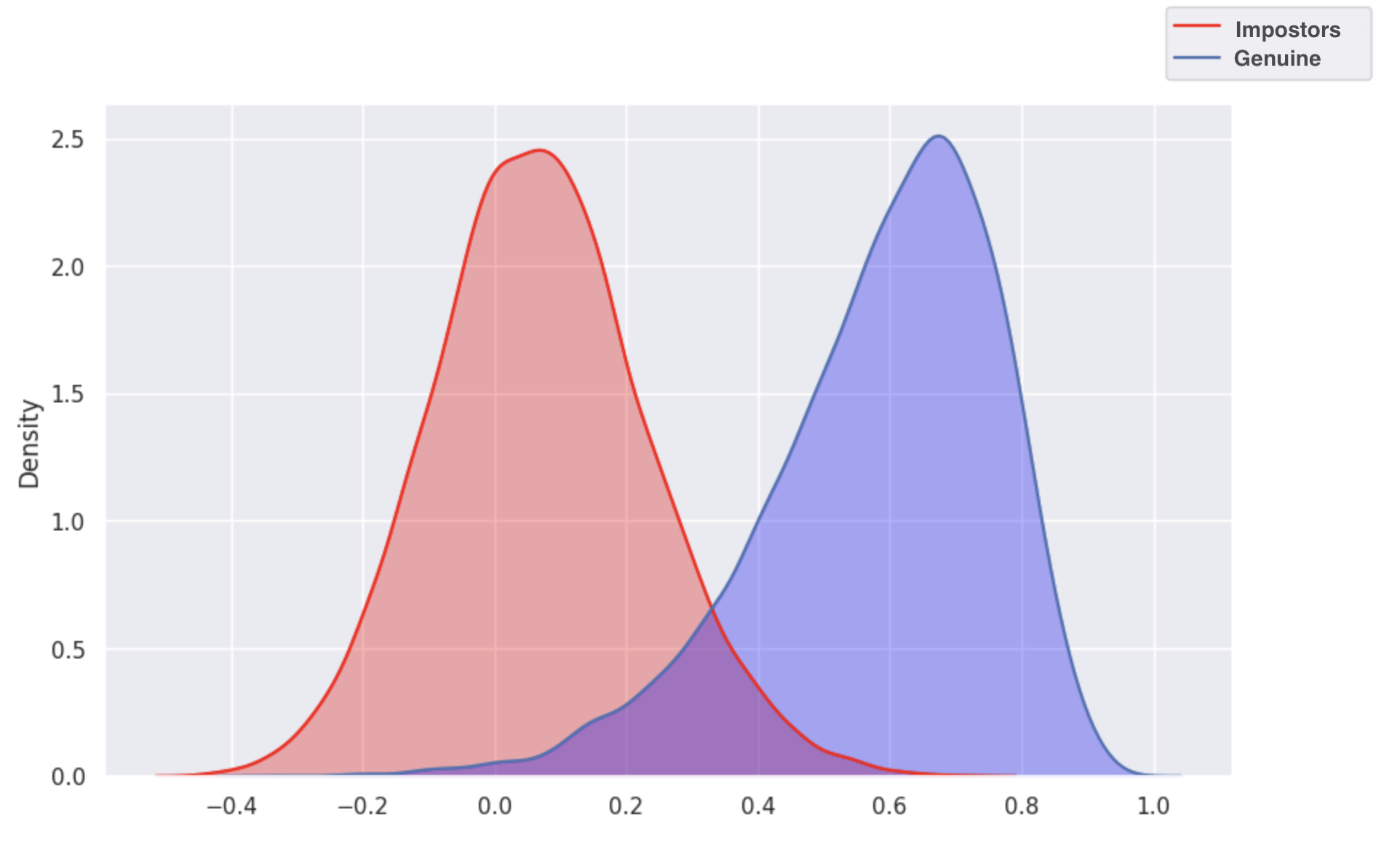}\\
    
    \subfloat[14x14]{\includegraphics[width=0.32\linewidth]{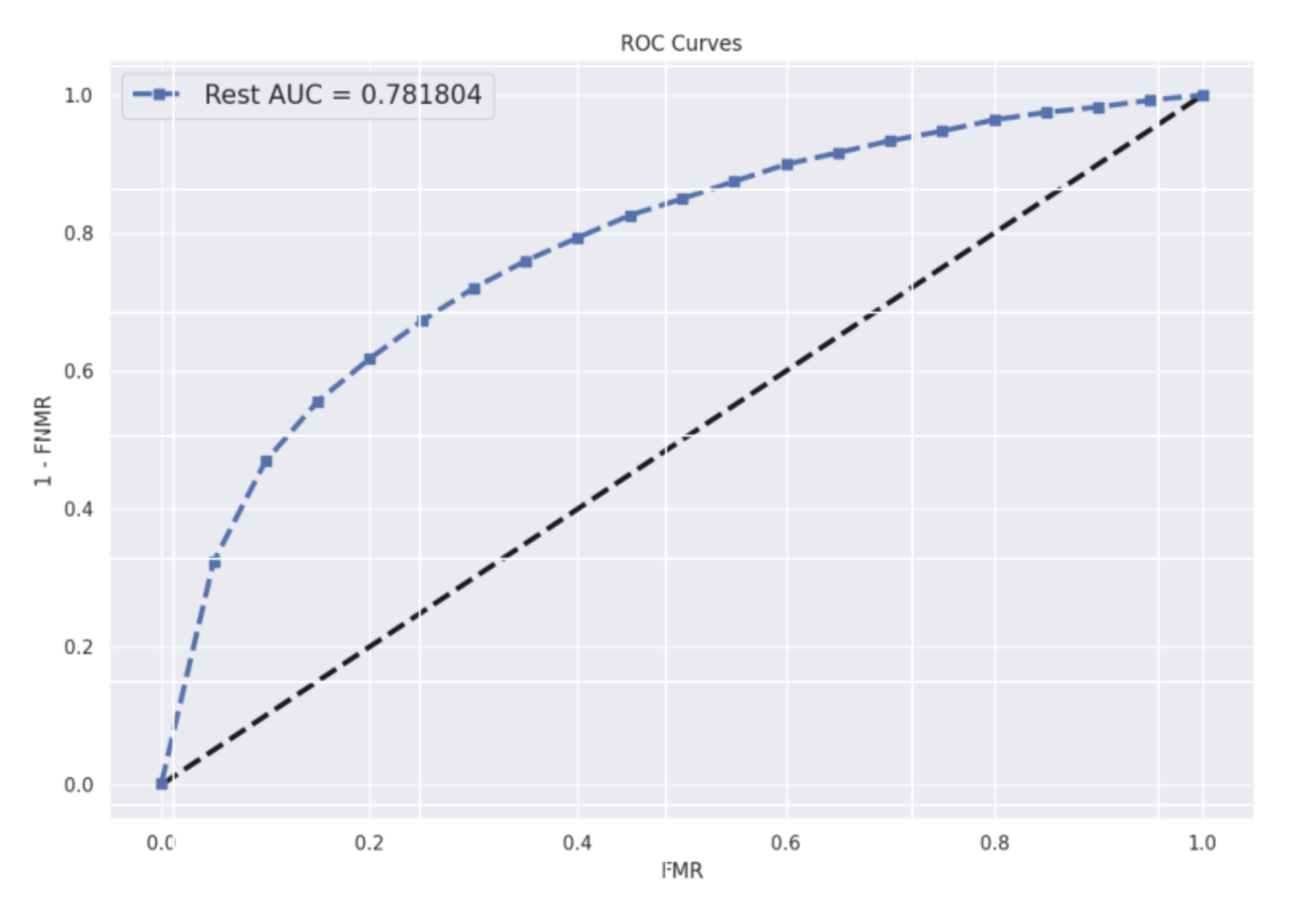}}
    \subfloat[28x28]{\includegraphics[width=0.32\linewidth]{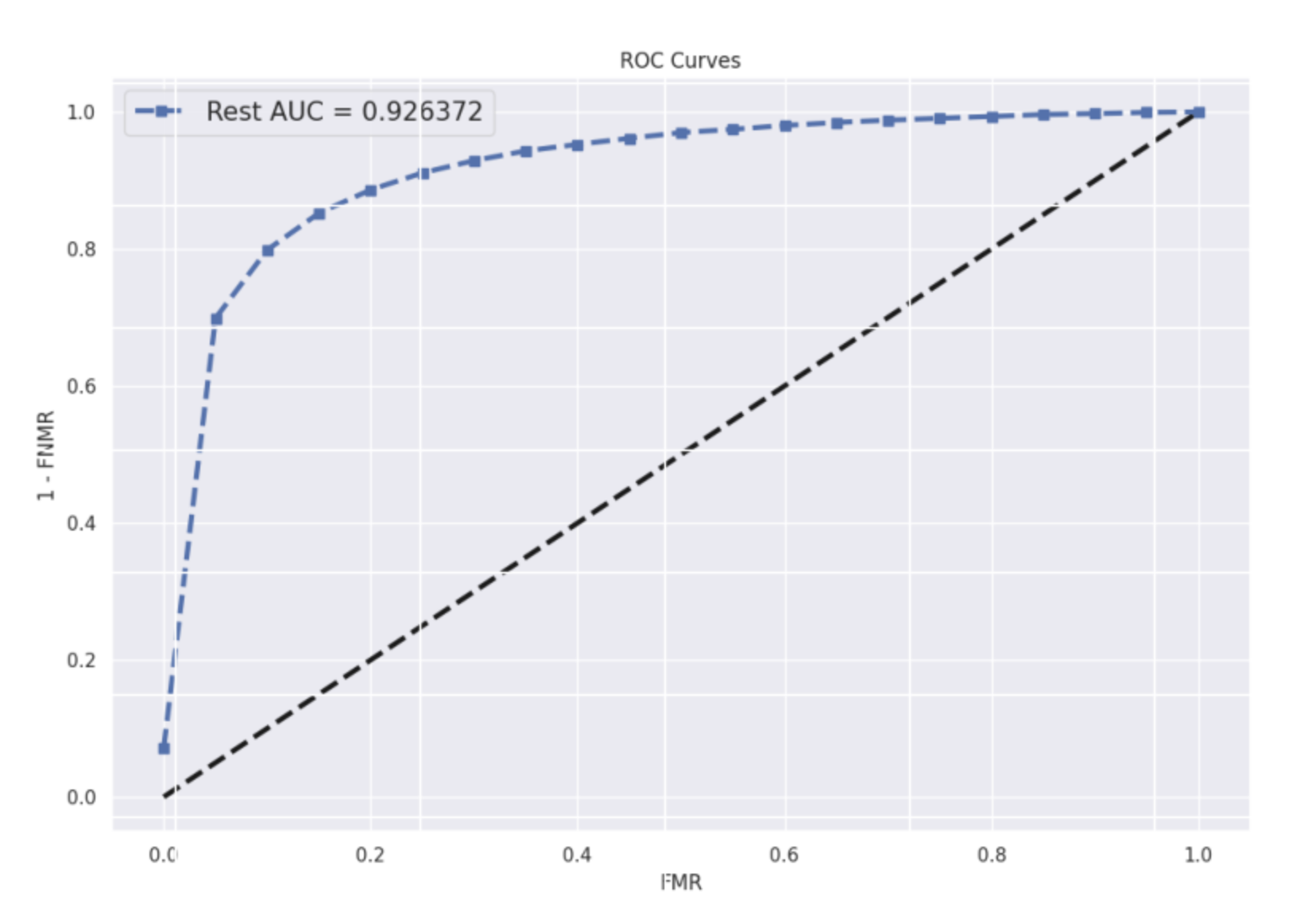}}
    \subfloat[56x56]{\includegraphics[width=0.32\linewidth]{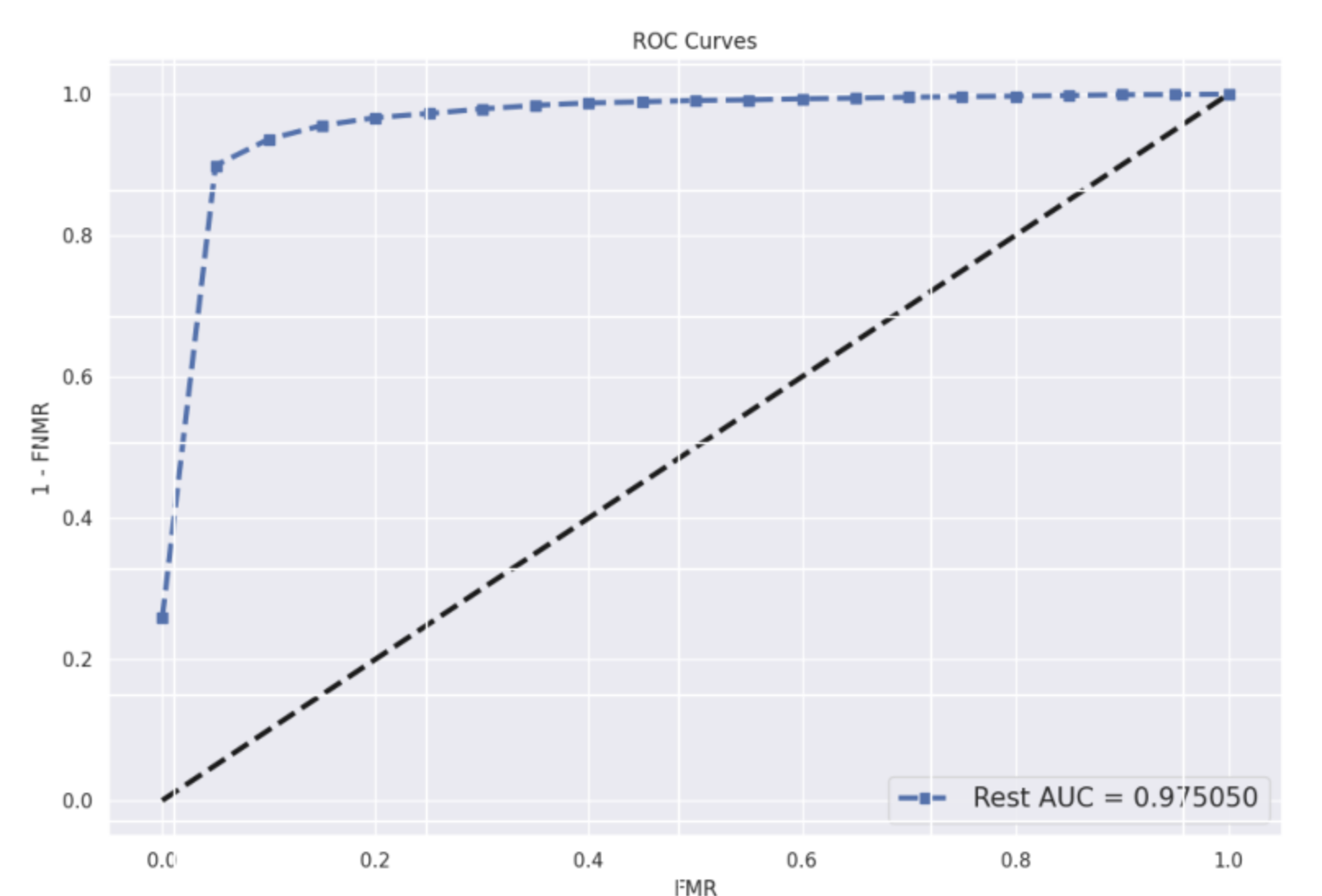}}
    
    \caption{Comparative image of the genuine and impostor curves and the ROC curves of the FTLGAN model on the test low resolution sets.}
    \label{curvasrocs}
    
\end{figure*}

The genuine and impostor curves, present in Figure \ref{curvasrocs} show how the separation between them improves as the resolution increases. An almost total separation in images of $56\times56$ pixels is particularly noticeable. Also, when analyzing the curves, it is evident that as the resolution increases, it is the genuine curve that mainly experiences an increase in the average distance between the pairs, going from an average distance of $0.3$ at $14\times14$ resolutions to an average distance of $0.65$ at $56\times56$. On the other hand, the impostor curve remains fixed close to a mean of $0$, which shows that the model is robust in detecting impostor pairs even at low resolution, but has difficulties in identifying genuine pairs at very low resolutions.

\begin{figure*}
    \centering
    \includegraphics[width=0.6\linewidth]{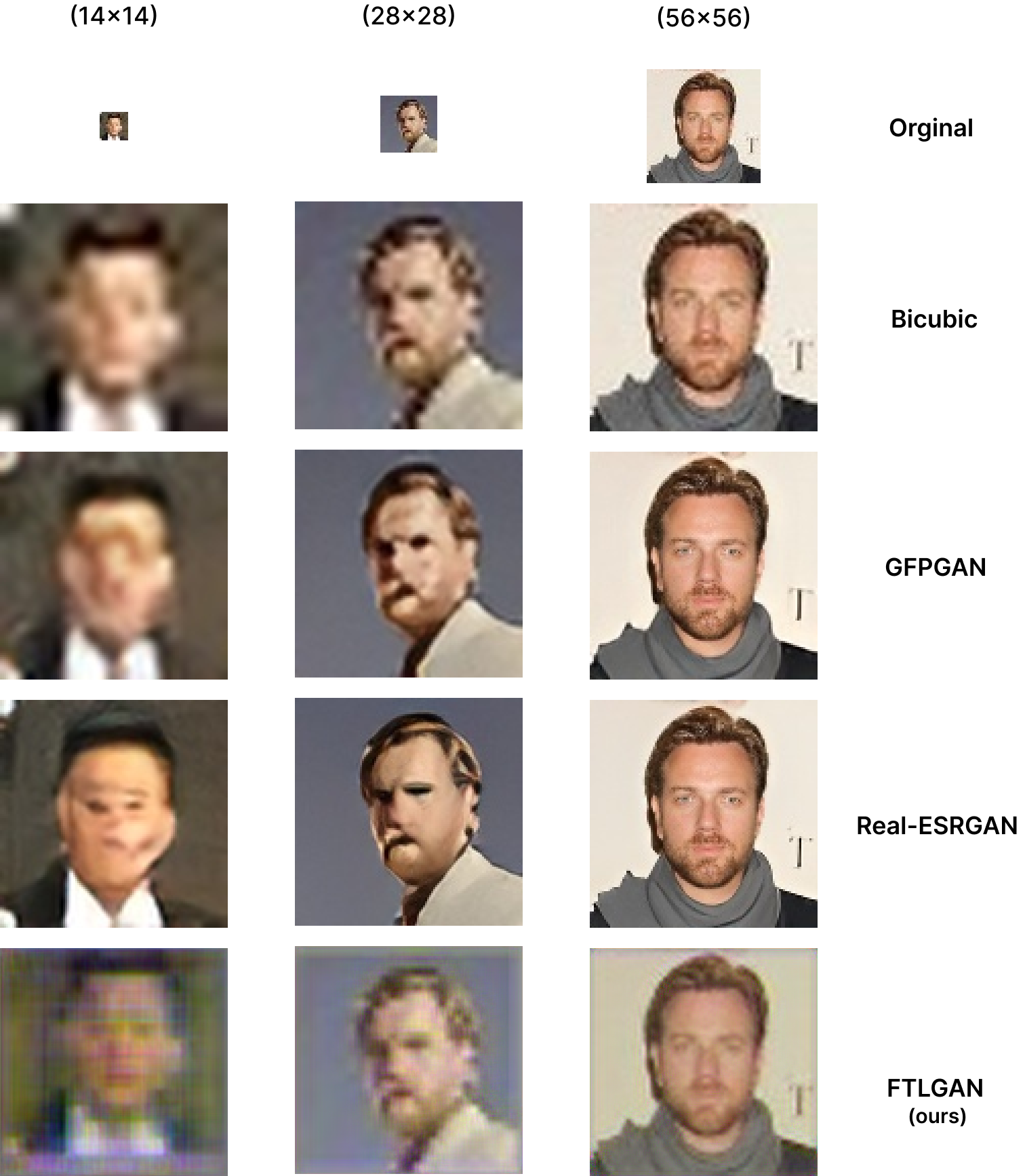}\\
    
    \caption{Visual results showing $14\times14$, $28\times28$ and $56\times56$ pixel images of actor Ewan McGregor processed by the bicubic, GFPGAN, Real-SRGAN, and FTLGAN models. The output of each model corresponds to a $112\times112$ pixel image.}
    \label{caras_comp}
    
\end{figure*}

In order to understand the results obtained, a visual SR test was performed with a recognizable public figure, Ewan Mcgregor, enlarging images in the three resolutions worked ($14\times14$, $28\times28$, and $56\times56$). Figure \ref{caras_comp} shows the results of four of the most important models: bicubic interpolation, GFPGAN, and Real-SRGAN compared with FTLGAN.

When observing the results of the images, it can be seen how the FTLGAN model shows similar results to those generated by a bicubic interpolation, but with subtle differences in the smoothness. It is important to note that before this study, interpolations offered the best results in this type of problem, and it is in this line that the FTLGAN model follows, by introducing less new information and taking better advantage of the visual information available, being a learning based model but behaving similarly to an interpolation model.

In contrast, it can be noted how the rest of the state-of-the-art learning based models tend to generate visually smoother results. Altering the identity of individuals, as occurs in $56\times56$ pixel images, or definitely deforming the face completely, as in the case of $28\times28$ and $14\times14$ resolutions, where the details of the identity are completely lost.

\subsection{Ablation study}

In this section, the impact of different elements on the FTLGAN model will be thoroughly analyzed through a series of experiments. Among the tests performed, the impact of using only real images in training versus synthetic images will be evaluated and the FR Angular ArcFace model, recognized for its accuracy in identifying facial features, will be used. In addition, several loss techniques will be integrated, including the well-known Mean Squared Error (MSE) along with the online triplet mining method \cite{Sikaroudi_2020}, which generates dynamic data triplets (anchor, positive and negative) during training by selecting difficult triplets from similar anchor and positive samples. These combined elements will provide a deeper understanding of the model performance in the super-resolution process. Each experiment will be performed on images exclusively at $28\times28$ pixels, an intermediate resolution that allows for agile evaluation and faithfully represents behavior between $14\times14$ pixels and $56\times56$ pixels. All combinations of experiments can be visualized in Table \ref{tabla00002}.

\begin{table}[!t]
\caption{Table of ablation experiments performed, the base experiment corresponds to the original FTLGAN model while the 5 subsequent experiments present variations in some point of the architecture that are highlighted in bold.}
\centering
\resizebox{0.42 \textwidth}{!}{ 
\begin{tabular}{|c|c|c|c|}
\hline \textbf{Exp} & \textbf{Type of images} & \textbf{Losses} & \textbf{Feat. extraction} \\
\hline \textbf{base} & Real & 
TL+ Precep & FaceNet \\
\hline \textbf{1} & \textbf{Synthetic} & 
\textbf{TL} & FaceNet \\
\hline \textbf{2} & Real & 
\textbf{TL} & FaceNet \\
\hline \textbf{3} & Real & \textbf{TL + MSE} & FaceNet\\
\hline \textbf{4} & Real & \textbf{OTL + Percep} & FaceNet\\
\hline \textbf{5 }& Real & TL + Percep & \textbf{Arcface} \\
\hline
\end{tabular}}

\label{tabla00002}
\end{table}

The results of the ablation study are presented in Table \ref{tabla_ablación}. Comparing the results of experiments 1 and 2, one can observe the positive impact of incorporating real images, resulting in a slight improvement of the $d'$ value from $2.036$ to $2.098$. These results indicate marginal improvements when using this type of images.

Similarly, when comparing Experiments 3 and 4 with the base experiment, it is observed that neither the use of online triplet mining nor the use of MSE loss generates a positive contribution in improving the model, reducing the $d'$ value to $2.017$ and $2.064$ respectively.

In the case of experiment 5, it is possible to notice a much lower performance than that obtained in the previous experiments with a $d' = 0.108$ because the model presented a divergent behavior during training. This behavior was presented in all the training that included an angular type face recognition model such as AdaFace \cite{ada} or CosFace \cite{wang2018cosface}. The details of this behavior will be addressed in the \ref{explicacion} section.

\section{Discussion of results} \label{explicacion}

\subsection{Why is FTLGAN the best model in the experiments?}

As can be seen in Figure \ref{caras_comp}, the results of learning-based models, such as GFPGAN, present smoother results compared to those delivered by the FTLGAN model. The latter seems to have a behavior closer to a bicubic interpolation. However, despite this, the FTLGAN results are superior at all resolutions, improving the $d'$. This effect may be due to several factors, the main one being the incorporation of face recognition embedding as part of the loss function.

\begin{table}[!t]
\caption{Comparative table of the results of the ablation experiments shown in Table \ref{tabla00002}}
\centering
\resizebox{0.18 \textwidth}{!}{ 
\begin{tabular}{|c|c|c|}
\hline \textbf{Exp} & \textbf{d'} & \textbf{AUC} \\
\hline \textbf{base} & \textbf{2.112} & \textbf{0.92} \\
\hline \textbf{1} & 2.036 & 0.92\\
\hline \textbf{2} & 2.089 & 0.91\\
\hline \textbf{3} & 2.064 & 0.92\\
\hline \textbf{4} & 2.017 & 0.91\\
\hline \textbf{5 }& 0.108 & 0.52\\
\hline
\end{tabular}}

\label{tabla_ablación}
\end{table}

The inclusion of face recognition embedding in the FTLGAN loss function incorporates the quality of face recognition into the face restoration, shifting the focus from image space, where models generally work, to the space of face representations. This shift in focus allows FTLGAN to use the limited information available in low-resolution faces to generate images that are more faithful to the original data.

In contrast, other GAN models tend to invent a lot of new information in order to smooth the image. These models often produce images that, while visually pleasing, may depart significantly from the original information contained in the low-resolution image. For example, in cases of $14\times14$ pixel images, where the information is contained in only 196 pixels, FTLGAN takes full advantage of this limited information to generate more accurate images by generating new pixels similar but not the same as those generated in interpolation. This ability of FTLGAN to maintain fidelity to the little information available is what allows it to outperform other models in terms of quality and accuracy in facial image restoration.

\subsection{Why does FaceNet work better than models like Arcface or Adaface?}

Angular loss-based models such as ArcFace and AdaFace have dominated the state of the art in the last 5 years in HR datasets such as LFW \cite{LFW}, however, as can be seen in Table \ref{tab:final} and Table \ref{tabla_ablación} angular models presented worse performances than Euclidean models in low-resolution cases.

The presented results can have several explanations, however, the main one is centered on the major problems that FR models are based on angular losses present in images with excessive noise or compression, which has been previously studied by in \cite{wang2018devil}. The results of the study show how the performance of angular models decays strongly as the image is damaged, an effect that is not as clearly seen in non-angular or contrastive model, those that perform best in these situations.

\subsection{Why does FTLGAN not converge with FR models based on angular losses?}

As visualized in \ref{tabla_ablación} the versions of FTLGAN based on angular losses do not converge, however, this fact was not an isolated case, since, when using optimizers other than SGD or when making changes in the learning rate, the model presented the same divergent and erratic behavior. These cases show that FTLGAN is a highly unstable model, which may be one of the major triggers of the instability seen when using models such as ArcFace or AdaFace.

This instability makes even more sense with the results visualized in Table \ref{tab:final} where it is possible to observe the poor performance of the angular models for classifying low-resolution images. This poor performance is likely to affect the stability of FTLGAN since the training of the generator is directly dependent on the face recognition model.

The instability of the model and the fact that FTLGAN converges only with certain specific parameters may be largely due to the fact that the model uses a Triplet Loss, which has been considered a highly unstable type of loss in the training process by numerous authors \cite{Wang2021}, so it is an important future task to improve the stability and convergence of this model in the future.

\section{Conclusions}

In recent years, advances in facial recognition have made this technique the most widely used biometric method, however, the inherent hardware problems still generate numerous cases in which the facial images obtained are of low resolution, which generates a strong loss of performance of the facial recognition models. Numerous solutions such as super-resolution have sought to improve the performance of these cases, however, the problems have persisted over time. Due to these problems of face recognition, the present work aimed to define the current limits of FR and propose a new solution to this problem, using the quality of face recognition as a training loss of SR models.

The work showed the poor performance of current super-resolution models, which focus on generating smoother, lifelike images, but in reality, perform poorly when it comes to face recognition. These poor results are largely due to the fact that traditional generative models do not incorporate face recognition as a primary task, making it a second-order objective.

Due to the low performance of face recognition at low resolution, this work developed a new super-resolution model: FTLGAN, which incorporates the quality of face recognition as a training loss using a triplet loss logic. This approach allows the development of a SR model focused on the quality of face recognition rather than the aesthetic quality of the image. The results of this model show a $d'$ 21\% higher than the best models of the current state of the art, specifically achieving a $d' = 1.099$ and $AUC = 0.78$ for $14\times14$ pixels, $d' = 2.112$ and $AUC = 0.92$ for $28\times28$ pixels, and $d' = 3.049$ and $AUC = 0.98$ for $56\times56$ pixels.

The positive results observed can be further explained by a detailed analysis of the two key contributions. First, by using real images for training, the model's performance improved, increasing the d' from 2.036 to 2.098 for $28\times28$ pixels. However, this improvement is marginal compared to the significant enhancement achieved by incorporating facial recognition embedding into the loss function. This latter approach raised the d' from 1.715 to 2.036 for $28\times28$ pixels. These findings underscore the importance of integrating facial recognition quality into the model’s training process for more effective low-resolution facial restoration, aligning with the promising results demonstrated by FTLGAN.

The development of this work opens a new line of research for future projects, allowing possible improvements in various face recognition problems such as image degradations with blurring or noise or even in recognition tasks with age changes, allowing real improvements in these tasks.

\section{Acknowledgment}
Fondecyt-Chile 1191131 and National Center for Artificial Intelligence CENIA FB210017, Basal ANID, partly supported this work.



\end{document}